%% file: authorsample.tex
\documentclass[graybox]{svmult}

\usepackage{type1cm}        
\usepackage{makeidx}         
\usepackage{graphicx}        
\usepackage{multicol}        
\usepackage[bottom]{footmisc}

\usepackage{listings}

\usepackage{url}
\usepackage[hidelinks,breaklinks]{hyperref}
\usepackage{amsmath}
\usepackage{float}
\usepackage{booktabs}
\usepackage{tikz}
\usepackage{multicol}
\usepackage{multirow}
\usepackage{rotating}
\usepackage{array}
\usepackage{color}
\usepackage{comment}
\usepackage{newtxtext}       %
\usepackage[varvw]{newtxmath}       

\renewcommand{\b}[1]{{\color{red} {\textbf{#1}}}}
\renewcommand{\a}[1]{{{\textbf{#1}}}}

\makeindex             

\begin{document}

\title*{On Representation of 3D Rotation in the Context of Deep Learning}
\author{Viktória Pravdová \orcidID{0009-0008-4074-1053}
\\ 
Lukáš Gajdošech\orcidID{0000-0002-8646-2147} \\ 
Hassan Ali\orcidID{0000-0001-9907-1834} \\ 
Viktor Kocur\orcidID{0000-0001-8752-2685} \\
}
\institute{Viktória Pravdová \at Faculty of Mathematics, Physics and Informatics, Comenius University 
\and Lukáš Gajdošech, \email{lukas.gajdosech@fmph.uniba.sk} \at Faculty of Mathematics, Physics and Informatics, Comenius University, 
\and Hassan Ali, \email{hassan.ali@uni-hamburg.de} \at Knowledge Technology, Department of Informatics, University of Hamburg, 
\and Viktor Kocur, \email{viktor.kocur@fmph.uniba.sk} \at Faculty of Mathematics, Physics and Informatics, Comenius University}
%
%
\maketitle
\vspace{-5mm}
\abstract{This paper investigates various methods of representing 3D rotations and their impact on the learning process of deep neural networks. We evaluated the performance of ResNet18 networks for 3D rotation estimation using several rotation representations and loss functions on both synthetic and real data. The real datasets contained 3D scans of industrial bins, while the synthetic datasets included views of a simple asymmetric object rendered under different rotations. On synthetic data, we also assessed the effects of different rotation distributions within the training and test sets, as well as the impact of the object's texture. In line with previous research, we found that networks using the continuous 5D and 6D representations performed better than the discontinuous ones.}
\input chapters/00-introduction.tex 
\input chapters/20-representations.tex
\input chapters/30-generation.tex
\input chapters/40-implementation.tex

\input chapters/50-results.tex

\input chapters/60-boxes.tex
\input chapters/90-conclusion.tex

\section*{Acknowledgments}

The work presented in this paper was carried out in the framework of the TERAIS project, a Horizon-Widera-2021 program of the European Union under the Grant agreement number 101079338. The results were obtained using the computational resources procured in the project National competence centre for high performance computing (project code:~311070AKF2) funded by European Regional Development Fund, EU Structural Funds Informatization of society, Operational Program Integrated Infrastructure.

\bibliographystyle{plain}
\bibliography{literature} 

\end{document}

%% file: chapters/00-introduction.tex
\section{Introduction}

Understanding and effectively representing 3D rotations is of fundamental importance in many areas of computer vision and robotics. Each rotation representation has its own application and limitations~\cite{Kim2023}, making the choice of representation a crucial factor in terms of the system's performance. In the context of deep learning, proper representation of rotations is a key factor for neural network accuracy and robustness in various tasks such as 3D object detection~\cite{Hodan2016}, pose estimation~\cite{Chen2021}, or 3D scene reconstruction~\cite{Jianwei2022}.

Deep neural networks 
have become a dominant approach in the field of 
machine learning in recent years. Thanks to their ability to extract complex patterns and features from large datasets, neural networks have achieved revolutionary results in various image processing, speech recognition, or natural language tasks. One of the most successful neural network architectures for image data processing are convolutional neural networks (CNN). These networks allow to efficiently capture spatial patterns and visual features at different levels of abstraction~\cite{Alzubaidi2021}.

In this work, we focus on the use of a residual neural network (ResNet)~\cite{resnet}, which is one of the leading CNN architectures. 
The ResNet architecture has demonstrated excellent results in many image processing tasks and has become a popular basis for further modifications and enhancements.

Traditional approaches to the representation of 3D rotations such as rotation matrices, Euler angles, or quaternions can lead to problems such as ambiguity in the representation or undesired discontinuities in the rotation representation~\cite{rot3d}. These shortcomings can negatively affect the learning process of neural networks and reduce their accuracy and robustness. There are several alternative approaches to the representation of 3D rotations that attempt to overcome the limitations of traditional methods by relying on an overparametrized representation. In addition to the rotation representation itself, the choice of the loss function used during training can also affect network performance. Not only that the rotation task comprises a variety of possible loss functions~\cite{Peretroukhin2020}, but it also requires consideration of the rotation properties to optimize the learning process~\cite{Hartley2012}.

In this paper, we evaluate the effects of different representations and loss functions on the accuracy and robustness of the trained networks. We also evaluate the effects of different training distributions and textures of input images. In order to control these factors we perform the evaluation on synthetically generated data containing an object rendered under varying rotations. We also validate the results on a real dataset.
In line with previous research~\cite{rot3d}, our results show that continuous representations result in more accurate and robust neural networks while also having more favorable training dynamics.

%% file: chapters/20-representations.tex
\section{Rotation Representations}
\label{chapter:representations} 

In this section, we will introduce the different possible representations of rotations in 3D and the concept of representation continuity.

3D rotations can be represented in different ways, which differ not only in their principle, but also in their dimensionality. It is important to note that a rotation in 3D space expressed by any representation is still just a mapping from 3D space to 3D space. Therefore, while general 3D rotations will always have 3 degrees of freedom, practical implementations of different representations may use more than 3 variables to describe them. Therefore, we define an $n-$dimensional representation as one which requires $n$ variables.


\subsection{Representation Continuity}

\label{sec:continuity}

The continuity of rotation representations has been shown to be important for their performance within deep neural networks, resulting in decreased errors \cite{rot3d}. We will test this hypothesis in this paper on our custom dataset (see Section \nameref{section:results}). 

We will now briefly introduce the intuitive reasoning for why this is the case by demonstrating the principles in a simper case of rotation in 2D space.

\begin{figure}
    \centering
    \includegraphics[width=8cm]{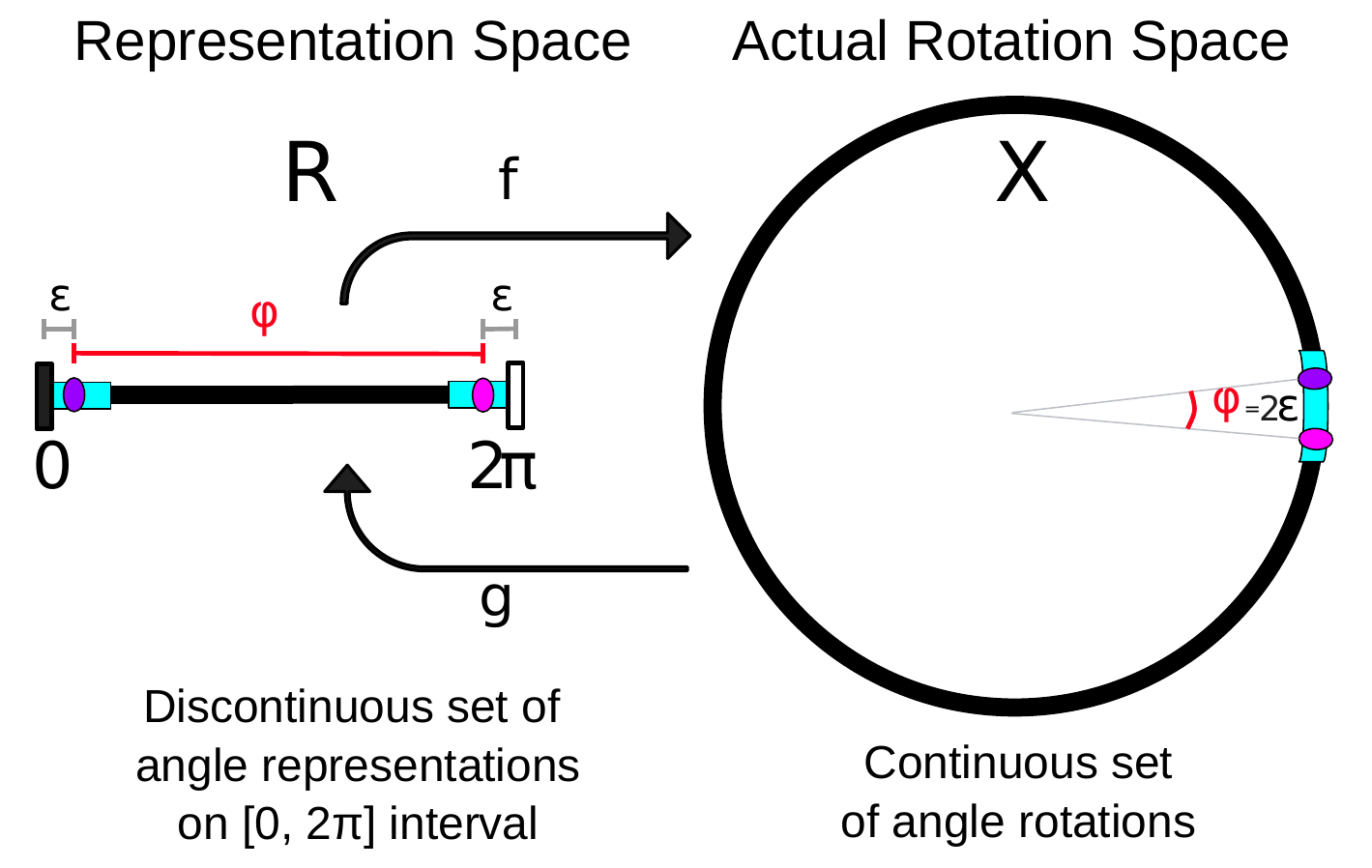}
    \caption[Illustration of the definition of a continuous representation on a 2D space.]{Illustration of the definition of a continuous representation on a 2D space. The region of discontinuity is shown in turquoise. In red is illustrated the problem when the size of the angle $\varphi$ differs in reality and in the representation: in reality it is small, but in the representation it is large.}
    \label{fig:continuity_definition}
\end{figure}

Consider the point $(1, 0)$. The set of images of this point under all 2D rotations about the point $(0, 0)$ (counterclockwise) will form a circle. Let us consider the space of rotations as this circle. We can then consider the representation space of 2D rotation as a line segment, or an interval between $0$ and $2\pi$, where the value represents the rotation angle. Both the rotation space and the representation space are shown in Figure~\ref{fig:continuity_definition}.

The conversions between the two spaces are intuitively obvious. However, a problem arises when the angle of rotation is close to $0$ and $2\pi$, respectively. Specifically, the angle between the purple and pink points in the figure would actually only be $2\epsilon$, but in the space of representations, the purple point would be given the value $\epsilon$ and the pink point $2\pi - \epsilon$. This is an example of discontinuous representation.

It is understandable why the demonstrated situation would be problematic while training neural networks. If we were to use a representation that returns large errors at rotations that are actually very close to the desired values, the neural network would continually retrain to different and more incorrect values. 

More rigorous definitions and theorems can be found in \cite{rot3d}, which also shows that such discontinuities occur for all representations of 3D rotation with dimension smaller than 5.

\subsection{Representations of 3D Rotation}
In this section we will introduce the different representations we will evaluate in Section~\ref{section:results}. We denote the dimensionality of the given representation in the subsection name.

A \textbf{rotation} is a geometric mapping that transforms points in space so that they rotate around a specified axis or point. In 2D space, the rotation is performed about a point, called the center of rotation, and the change in position of the points is determined by the magnitude of the angle of rotation. In 3D space, rotation can be performed about an axis, which means that the points rotate in any plane perpendicular to that axis as if 2D rotation was performed.

\subsubsection{Rotation matrices -- 9D} 
\label{cap:rotmat}
The most straightforward representation of rotations are rotation matrices, since they can be directly used for point transformations via matrix-vector multiplication. Each 3D rotation can be represented as an orthogonal matrix $R$ with $\text{det}(R) = 1.$ This representation also makes it easy to show that 3D rotations form the special orthogonal group $SO(3)$ and can thus be composed and inverted via matrix multiplication and inversion (transposition in case of orthogonal matrices).

Even though this representation is continuous~\cite{rot3d}, rotation matrices are unsuitable for use in neural networks. Using them on output of networks would yield 9 output parameters and it would be necessary to ensure that the resulting matrix belongs to $SO(3)$, e.g. that the columns are mutually orthonormal and $\text{det}(R) = 1$, which is difficult to ensure and failing to do so may result in \textit{deformed} matrices that will not only perform rotation, but also other transformations such distortion, magnification, projection, or other unexpected behaviors. 



\subsubsection{Euler angles -- 3D}
One of the most compact and simplest representations of 3D rotation is the representation using Euler angles. 
Each 3D rotation can be composed of three successive applied rotations about the individual axes proper to the object. There are a number of conventions for the order in which the axes an object is rotated are chosen, but we will consider that rotations are performed sequentially around the $x$, $y$ and $z$ axes.

The representation space for a particular ordering of axes is a subset of $\mathbb{R}^3$, where the vector elements represent the three angles.

The rotation by Euler angles can be thought of as a composite representation, so it is the product of three matrices, each with one parameter -- the angle of rotation about a given axis: 
\begin{equation} 
\scriptsize
Z(\alpha) Y(\beta) X(\gamma) = 
    \begin{bmatrix}
    \cos\alpha \cos\beta & \cos\alpha \sin\beta \sin\gamma - \cos\gamma \sin\alpha & \sin\alpha \sin\gamma + \cos\alpha \cos\gamma \sin\beta \\
    \cos\beta \sin\alpha & \cos\alpha \cos\gamma + \sin\alpha \sin\beta \sin\gamma & \cos\gamma \sin\alpha \sin\beta - \cos\alpha \sin\gamma \\
    - \sin\beta & \cos\beta \sin\gamma & \cos\beta \cos\gamma 
    \label{eq:euler2matrix}
   \end{bmatrix}.
\end{equation}
The problem with this representation is that there is no bijection between this space and the space $SO(3)$,
since for some matrices of $SO(3)$ there is an ambiguity when converting to angles. In addition, a \textit{gimbal lock} can occur, where the two axes align and a degree of freedom in the rotation is lost. Similar to the example presented in Section~\ref{sec:continuity} this representation is discontinuous.


\subsubsection{Axis-angle -- 3D $\And$ 4D
}
Every rotation in 3D space is equivalent to a rotation about some axis (vector), the so-called Euler axis, by a certain angle. Based on this fact it is possible to construct the space of Axis-angle representations as a subset of $\mathbb{R}^4$, where the first three components represent the axis (unit length) and the last is the angle. It is also possible to represent the angle by the norm of the axis vector. Such a representation is also called an Exponential representation, and its representation space is a subset of $\mathbb{R}^3$.

If $R$ is a rotation matrix, then we can obtain the axis as $\vec{u}$ using:
\begin{equation}
\begin{array}{cccc}
 \theta = \arccos \dfrac{\text{Tr}(R) - 1}{2}
&&&
 \vec{u} = \dfrac{\theta}{2\sin\theta} \begin{pmatrix}
 R_{32} - R_{23} \\ R_{13} - R_{31} \\ R_{21} - R_{12}
\end{pmatrix}
\end{array},
\label{eq:mat2axis}
\end{equation}
where $\theta$ is the angle and $\vec{u}$ is the axis such that $||\vec{u}|| = \theta$. Both of these representations are discontinuous~\cite{rot3d}. To obtain $R$ from this representation it is possible to use the Rodrigues' formula.



\subsubsection{Unit quaternions -- 4D}
Similarly, quaternions are based on the idea of an axis around which rotation takes place. Let the unit vector of this axis be $(x, y, z)$, and the angle we want to rotate by be $\theta$, then the quaternion representing this rotation is:
\begin{equation}
    \vec{q} = \left(\cos\frac{\theta}{2}, x\sin\frac{\theta}{2}, y\sin\frac{\theta}{2}, z\sin\frac{\theta}{2}\right).
\end{equation}
We see that the space of this representation is a subset of $\mathbb{R}^4$. In particular, since these are unit quaternions, their space is a $4$ dimensional unit hypersphere.


The image of a point $\vec{p}$ after rotation by a quaternion $\vec{q}$ is given by $\vec{p}\prime = \vec{q}\vec{p}\vec{q}^{-1}$, where $\vec{q}^{-1}$ is the associated unit quaternion of $q$ (it has a negative imaginary component with respect to $\vec{q}$). We can also express this rotation by the matrix $R$, which has the following form: 
\begin{equation}
 R = f_{quat}(\vec{q}) = \begin{bmatrix}
1 - 2 (q_j^2 + q_k^2) &
2 (q_i q_j - q_k q_r) &
2 (q_i q_k + q_j q_r) \\
2 (q_i q_j + q_k q_r) &
1 - 2 (q_i^2 + q_k^2) &
2 (q_j q_k - q_i q_r) \\
2 (q_i q_k - q_j q_r) &
2 (q_j q_k + q_i q_r) &
1 - 2 (q_i^2 + q_j^2)
\end{bmatrix}.
\label{eq:quat2mat}
\end{equation}


In this representation, ambiguity arises when the transformation is identical (i.e., when the degree of rotation of $\theta = 0$) or when rotated by $\pi$. This causes the discontinuity of this representation \cite{rot3d}.

\subsubsection{Gram-Schmidt -- 6D}
Let us return again to the idea of using the full rotation matrix for representation. We can represent matrices from $SO(3)$ by $9$ numbers, but these must satisfy the conditions of orthogonality and positive determinant. This means, that the matrix columns must be orthonormal, which can be achieved from arbitrary linearly independent vectors by the process of Gram-Schmidt orthonormalisation. In this process, the vectors are taken sequentially from the input, first adjusted to be orthogonal to all previously processed vectors, and then normalized.

The columns of each rotation matrix form an orthonormal basis of $\mathbb{R}^3$, so we only need $n-1$ (in our case $2$) columns to determine the last column easily. 
This brings us to the 6D representation, which for a change is continuous~\cite{rot3d}.

We can use the following procedure to convert from the space of representations ($\mathbb{R}^{3\times2}$) to the space of rotation matrices:
\begin{equation}
    f_{GS} \left( \begin{bmatrix}
        \vec{a}_1,   & \vec{a}_2     \\
    \end{bmatrix} \right)
    =
    \begin{bmatrix}
        \vec{b}_1,   & \vec{b}_2,   & \vec{b}_3  \\
    \end{bmatrix} 
    = 
    \begin{bmatrix}
        N(\vec{a}_1), &
        N\left(\vec{a}_2 - \left<\vec{b}_1 | \vec{a}_2\right>\vec{b}_j\right), &
        \vec{b}_1 \times \vec{b}_2
    \end{bmatrix},
    \label{eq:fGS}
\end{equation}
where $N(\vec{x}) = \vec{x} / ||\vec{x}||$ and $\times$ is the cross product which ensures that $\text{det}(R)=1$. For the reverse conversion, we simply drop the last column of the matrix. In further text we denote this representation as GS.

\vspace{-3mm}
\subsubsection{Stereographic projection -- 5D}
Following \cite{rot3d}, we also introduce a 5D representation, which is an enhancement of 6D GS representation by stereographic projection and normalization, which also preserves the continuity of this representation. 
To convert the rotation matrix $R \in \mathbb{R}^{3\times3}$ into the representation space $\mathbb{R}^5$ we use the function $g_P : \mathbb{R}^9 \to \mathbb{R}^5$ \cite{rot3d}:
\begin{equation}
    g_P( g_{GS}(R) ) = g_P(M) = (M_{11}, M_{21}, *P(M_{31}, M_{12}, M_{22}, M_{32})),
    \label{eq:gP}
\end{equation}
where $*$ denotes the element enumeration and $P$ is the  normalized stereographic projection~\cite{rot3d} that returns a $3$ element vector in this case.

To inversely convert the vector $\vec{u} \in \mathbb{R}^5$ to the space of rotation matrices, we in turn use the function $f_P : \mathbb{R}^5 \to \mathbb{R}^9$:
\begin{equation}
    f_P(\vec{u}) = f_{GS} ([u_1, u_2, *Q(u_3, u_4, u_5)] ^{3\times2}),
    \label{eq:fP}
\end{equation}
where $Q$ is the inverse normalized stereographic projection~\cite{rot3d}, which here returns a $4$ element vector. The superscript $3\times2$ denotes the rearrangement of the vector into a matrix, which is then the input to the GS inverse function of the $g_{GS}$ representation \eqref{eq:fGS}.

%% file: chapters/30-generation.tex
\section{Synthetic Data Generation}
\label{section:generation}

Training neural networks requires abundant data~\cite{Voynov2023}. In order to asses the effects of different characteristics of training data on the performance of the selected network we use synthetically generated data. This allows us to create datasets with different characteristics. To generate the data we render $256\times256$ image of the same object on a neutral background under different rotations sampled from different distributions.



\subsection{Object}

\begin{figure}[h]%
\vspace{-3em}
    \centering
    {{\includegraphics[width=35mm]{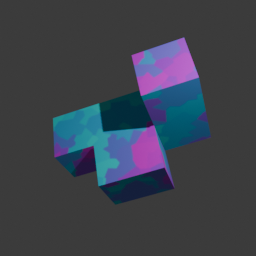} }}%
    {{\includegraphics[width=35mm]{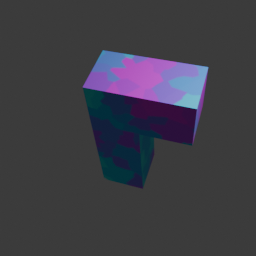} }}%
    {{\includegraphics[width=35mm]{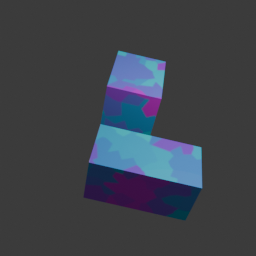} }}%
    \caption{Rendered images of the designed object under varying rotations.}%
    \label{fig:show_cool}%
\vspace{-1.5em}
\end{figure}

A suitable object should not contain any symmetries (since rotational ambiguity can occur with symmetries)~\cite{Pitteri2019}. We chose an object consisting of five cubes glued together with walls such that no two sides look the same. However, in order to avoid symmetries within the bases, we added texture. As a simple texture, we choose a coloured Voronoi mosaic. We used Blender in conjunction with the Python library \texttt{bpy}\footnote{https://docs.blender.org/api/current/index.html} to model and generate the images. The object rendered under different rotations is shown in Fig.~\ref{fig:show_cool}.

\subsection{Rotation Distributions}
We made several datasets with the same object, varying the rotation distributions of the training, validation and test sets of the dataset.

In order to define the dataset difficulty we will first define a rotation neighborhood. Consider some rotation matrix $R_i$ and an angle $\phi$. A rotation matrix $R$ belongs to the \textbf{rotation neighborhood $\rho(R_i, \phi)$} of a matrix $R_i$ if the angle between matrices $R$ and $R_i$ is less than $\phi$. Formally by \eqref{eq:angle_rotmat}: $R \in \rho(R_i, \phi) \equiv e_{RE}(R_i, M) < \phi$ .

Each distribution will be defined by a set of rotation matrices $R_i$ and an angle $\phi$. The test set will contain only the views for which $R \in \rho(R_i, \phi)$ for at least one $i$. The remaining views will be contained in the training and validation sets with an 80/20 split.

To visualize these distributions, we will represent the rotations using a sphere, where each rotation is represented as the position of the point $(1, 0, 0)$ after it is rotated. This representation misses one degree of freedom since the surface coordinates of the sphere have only two degrees of freedom, while the rotations have three, resulting in flattened shapes. Nevertheless, we find this visualization to be sufficient to illustrate the differences in the dataset distributions.


Each dataset contains $11\ 000$ data points in the form of images and matrices, with $1\ 000$ of images in the test set, $2\ 000$ in the validation set, and the remaining $8\ 000$ in the training set.
We also introduce the notion of a task difficulty metric, call it \textbf{median distance to training set}\label{def:zeta} and denote it $\zeta$. We define it as the median angle between matrices in the test set and matrices closest to them in the training set.

\subsubsection{Random Distribution -- $\zeta = 6.3^\circ$}
\begin{figure}[h]
    \vspace{-3em}
    \centering
    \includegraphics[width=46mm]{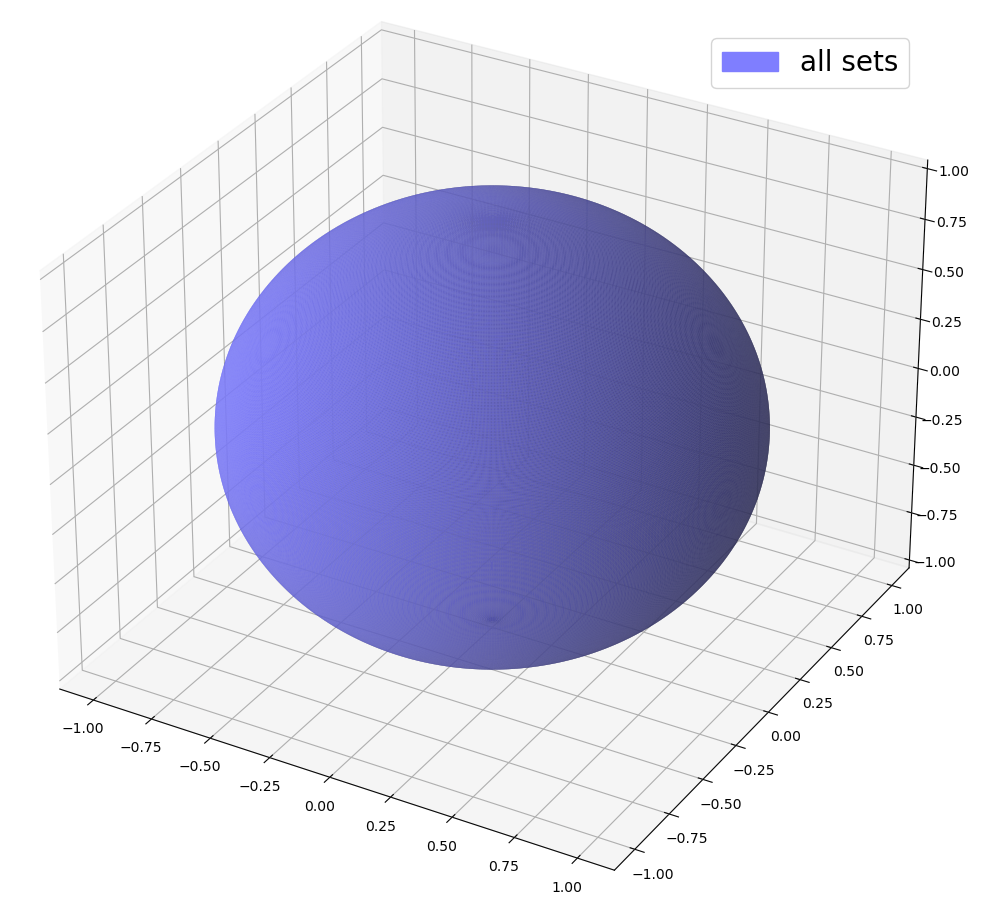}
    \includegraphics[width=46mm]{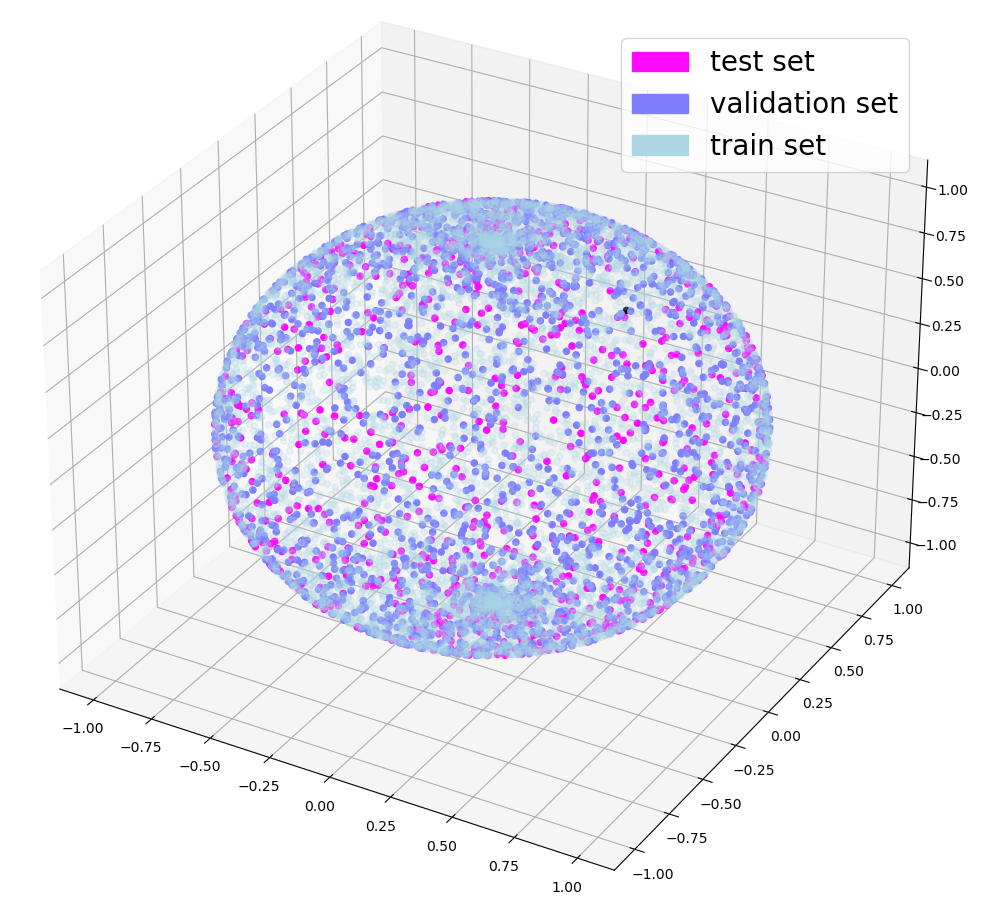}
    \caption[Random distribution.]{Random Distribution. Theoretical area on the left, actual samples in the dataset on the right.}
    \label{fig:random}
\end{figure}

For the first dataset, we generate the rotations of the training and test sets from the same distribution. We will use the space of all rotations by sampling random Euler angles from a uniform distribution on the interval $0^\circ$ to $360^\circ$ (see Fig.~\ref{fig:random}).


\subsubsection{Big Hole Distribution -- $\zeta = 17.5^\circ$}
\begin{figure}[h]
    \centering
    \vspace{-3em}
    \includegraphics[width=46mm]{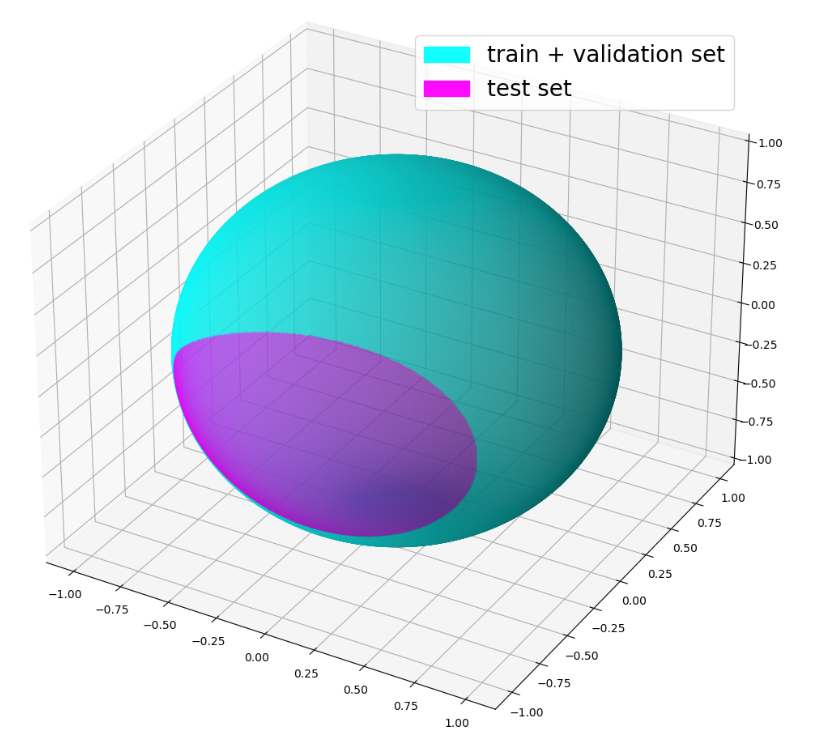}
    \includegraphics[width=46mm]{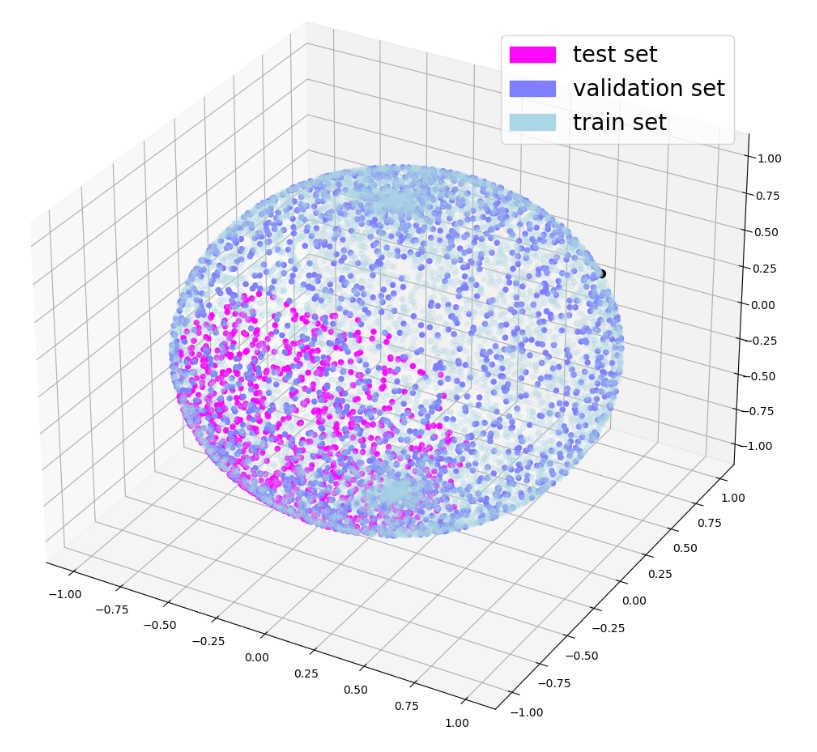}
    \caption[Distribution with a big hole.]{Big Hole Distribution. Theoretical area on the left, actual samples in dataset on the right.}
    \label{fig:big_hole}
    \vspace{-0.5em}
\end{figure}

For the second dataset, we randomly choose only one rotation matrix $R_0$ and an angle $\phi = 50^\circ$. This gives us one large rotation neighborhood for the test set (Fig.\ref{fig:big_hole}). This will allow us to evaluate the whether the neural network will be able to generalize to the larger unexplored region.

\subsubsection{Many Holes Distribution -- $\zeta = 17.5^\circ$}

\begin{figure}[h]
    \centering
    \includegraphics[width=46mm]{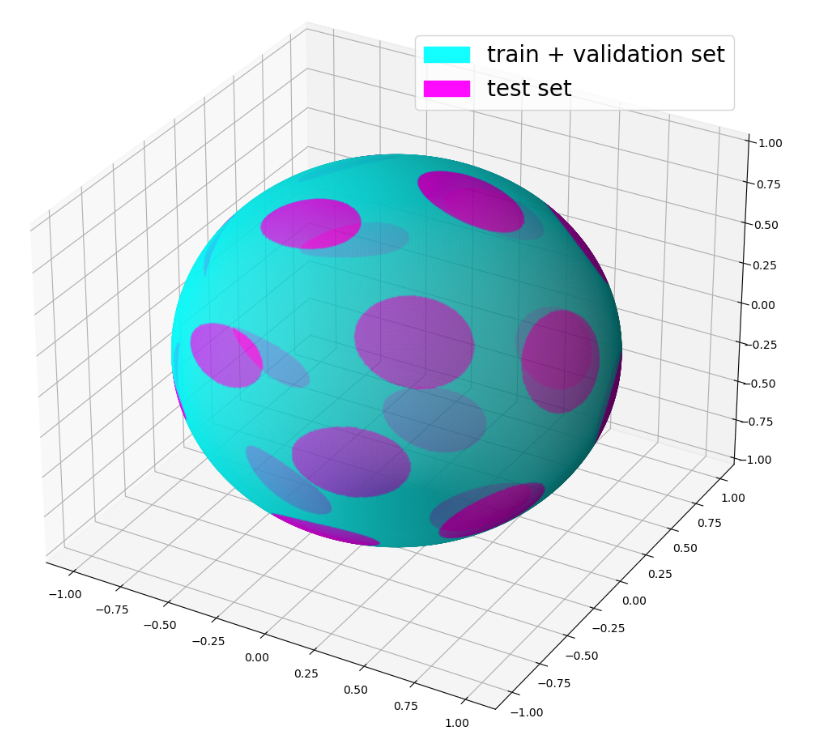}
    \includegraphics[width=46mm]{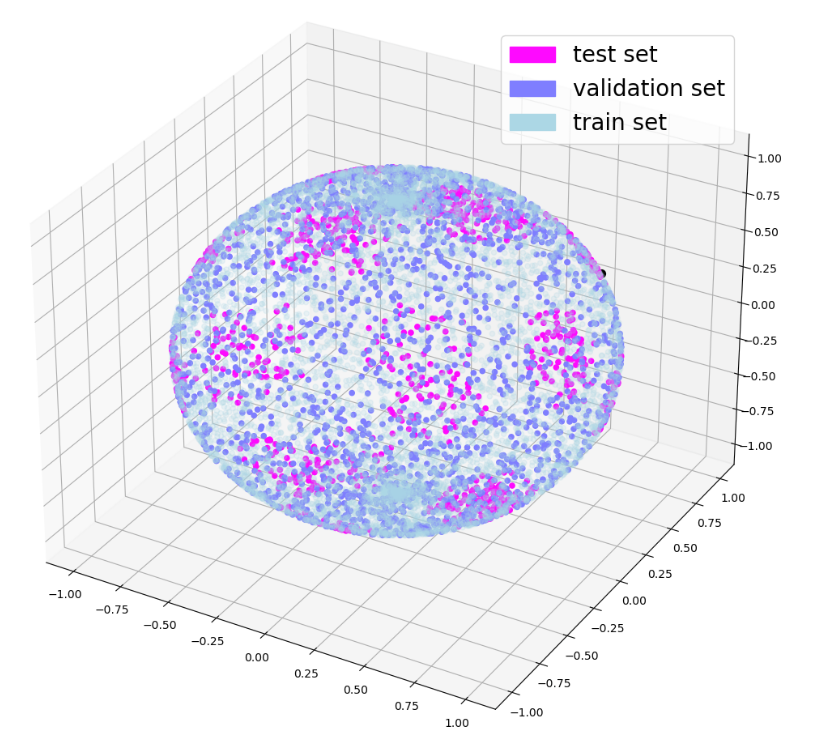}
    \caption[Distribution with many holes.]{Many Holes Distribution. Theoretical area on the left, actual samples in dataset on the right.}
    \label{fig:derava}
    \vspace{-1.5em}
\end{figure}

We defined this distribution with $20$ rotation matrices, each with the same angle $\phi = 15^\circ$ (see Fig. \ref{fig:derava}). To produce pseudo-regularly distributed rotation matrices, we used the fibonacci spiral~\cite{fibonaci_spiral}, which produced uniformly distributed points on the surface of the sphere. We then sampled for a rotation matrix that rotates a point (1, 0, 0) to a given point and used it.

\subsubsection{Colored versus Monochrome Texture -- $\zeta = 6.3^\circ$}
\begin{figure}[h]
    \centering
    {{\includegraphics[width=32mm]{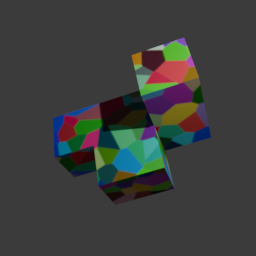} }}%
    {{\includegraphics[width=32mm]{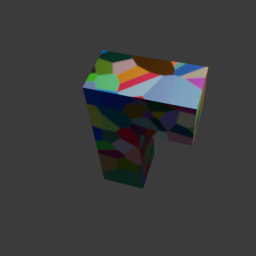} }}%
    {{\includegraphics[width=32mm]{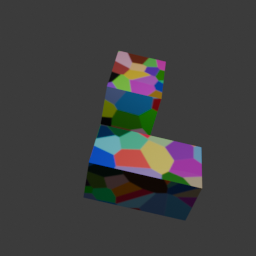} }}%
    \caption{Samples of the multicoloured object rendered under varying rotations.}%
    \label{fig:show_color}%
\end{figure}
\begin{figure}[h]
    \centering
    {{\includegraphics[width=32mm]{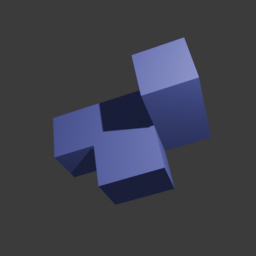} }}%
    {{\includegraphics[width=32mm]{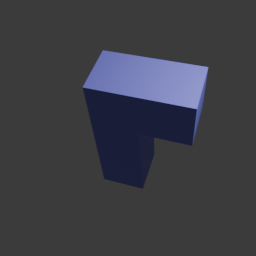} }}%
    {{\includegraphics[width=32mm]{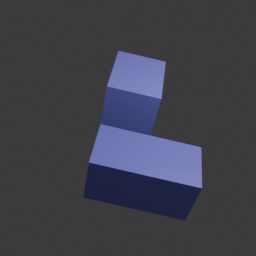} }}%
    \caption{Samples of the monochrome object rendered under varying rotations.}%
    \label{fig:show_one}%
    \vspace{-1.0em}
\end{figure}

We also decided to test the effect of the texture used on the object on the training results. Therefore, we also created two additional sets of images and matrices, one using a more strongly colored object (Fig.~\ref{fig:show_color}) and the other using a monochromatic object (Fig.~\ref{fig:show_one}). The shape, distribution, and even the exact rotations were the same as for the random distribution set.


%% file: chapters/40-implementation.tex
\section{Methodology}

In this section we describe the trained models in terms of their architecture and the loss functions used. 

\subsection{Architecture}


All of our models are based on the ResNet18~\cite{resnet} backbone with a linear layer after global pooling. When using regression losses, the output of the linear layer is the same size as the dimension of the representation. 
For the axis angle representation (denoted as A-A) we use two different parametrizations. In the 3D parametrization the angle of rotation is the norm of the vector and in the 4D parametrization the vector represents only the axis direction and the angle is represented by a separate scalar value.

We also created architectures in which we turn the problem of regressing angles into a classification problem. For each variable representing an angle, we created $180$ or $360$ categories, each representing a range of one degree.

We used this approach in the \texttt{Euler bin} model where we use linear output layer with softmax activation, but chose up to $360 \times 3$ neurons in the output, with the first $360$ belonging to the first angle and so on. In addition, we also made a combined model \texttt{A-A bin} where the axis was represented with a 3-dimensional vector, and the angle as a separate classification output with $180$ bins. We chose the size $180$ in this case because, unlike the previous case of Euler angles, here the orientation of the axis can be changed via the first parameter.

\subsection{Loss functions}

For regression models, we use three types of loss functions:

\begin{itemize}
\item Rotation error in degrees $e_{RE}$:
\begin{equation}
    e_{RE}\left(R, \hat{R}\right) = \arccos\left(\frac{\operatorname{tr}(R^{\top}\hat{R}) - 1}{2}\right),
    \label{eq:angle_rotmat}
\end{equation}%
where $R$ is the output rotation matrix and $\hat{R}$ is the ground truth rotation matrix~\cite{Hodan2016}. To calculate the rotation error in degrees we usually perform conversion to the matrix representation first.
\item Angle between the estimated vector and ground truth direction:
\begin{equation}
e_{TE}(\vec{y}, \vec{\hat{y}}) = \text{arccos}\left(\frac{\left< \vec{y} | \vec{\hat{y}}\right>}{||\vec{y}|| ||\vec{\hat{y}}||}\right). 
\label{eq:vector_angle}
\end{equation}
\item L2 loss function:
\begin{equation}
L(\vec{y}, \vec{\hat{y}}) = ||\vec{y} - \vec{\hat{y}}||^2,
\label{eqn:l2}
\end{equation}%
where $\vec{y}$ represents the output and $\vec{\hat{y}}$ represents the ground truth vector.
\end{itemize}

For classification outputs we use the cross-entropy which is defined for softmax activation as
\begin{equation}
    L = -\sum_{i = 1}^N y_i log(p_i)
    \label{eq:CE},
\end{equation}
where $y_i$ is an indicator variable with value 1 for the true class and 0 for all others, and $p_i$ is the predicted softmax probability for each of the $N$ classes.

We summarize the different model configurations in Table \ref{tab:implement}.

\newcolumntype{L}{>{\centering\arraybackslash}m{2cm}}
\newcolumntype{R}{>{\centering\arraybackslash}m{2cm}}
\begin{table}
    \centering
    \resizebox{\textwidth}{!}{
    \begin{tabular}{|c| R L p{8.5cm}|}
    \hline
    \textbf{Id} & \textbf{Representation} & \textbf{Loss function} & \textbf{Implementation Details} \\
\hline
1 & \texttt{Euler} & $e_{RE}$ (\ref{eq:angle_rotmat}) & 
Conversion of output to matrix representation using kornia. \\ \hline

2 & \texttt{Euler} & $L2$ (\ref{eqn:l2}) & 
Conversion of output to matrix representation using kornia.   \\ \hline

3 & \texttt{Euler  bin} & $e_{RE}$ (\ref{eq:angle_rotmat}) &  
The angles are obtained as weighted averages and then converted to matrix representation using kornia. \\ \hline

4 & \texttt{Euler  bin} & $CE$ (\ref{eq:CE}) & 
 \\ \hline

5 & \texttt{Quaternion} & $e_{RE}$ (\ref{eq:angle_rotmat}) & 
Conversion of output to matrix representation using kornia. \\ \hline

6 & \texttt{Quaternion} & $L2$ (\ref{eqn:l2}) & 
\\ \hline

7 & \texttt{A-A 3D} & $e_{RE}$ (\ref{eq:angle_rotmat}) & 
Conversion of output to matrix representation using kornia.  \\ \hline

8 & \texttt{A-A 3D} & $L2$ (\ref{eqn:l2}) & 
 \\ \hline

9 & \texttt{A-A 3D} &  $e_{TE}$ (\ref{eq:vector_angle}) + $L2$ (\ref{eqn:l2}) & 
Loss is the sum of $e_{TE}$ for the axis vector and $L2$ of its norm. \\ \hline

10 & \texttt{A-A 4D} & $e_{RE}$ (\ref{eq:angle_rotmat}) & 
Conversion of output to matrix representation using kornia. \\ \hline

11 & \texttt{A-A 4D} & $L2$ (\ref{eqn:l2}) & 
\\ \hline

12 & \texttt{A-A 4D} & $e_{TE}$ (\ref{eq:vector_angle}) + $L2$ (\ref{eqn:l2}) & 
$e_{TE}$ for the axis vector and $L2$ for the additional scalar. \\ \hline

13 & \texttt{A-A bin} & $CE$ (\ref{eq:CE}) + $L2$ (\ref{eqn:l2}) & 
Loss is the sum of $CE$ on the bin angle and $L2$ on the axis. \\ \hline

14 & \texttt{A-A bin} & $CE$ (\ref{eq:CE}) + $e_{TE}$ (\ref{eq:vector_angle}) &
Loss is the sum of $CE$ on the bin angle and $e_{TE}$ on the axis. \\ \hline

15 & \texttt{Stereo} & $e_{RE}$ (\ref{eq:angle_rotmat}) & 
Conversion of the output using \eqref{eq:fP} a \eqref{eq:fGS}.
\\ \hline

16 & \texttt{Stereo} & $L2 + L2$ (\ref{eqn:l2}) &
$L2$ on two vectors after conversion to GS representation (\ref{eq:fGS}). \\ \hline
 
17 & \texttt{Stereo} & $e_{TE} + e_{TE}$ (\ref{eq:vector_angle}) & 
$e_{TE}$ on two vectors after conversion to GS representation (\ref{eq:fGS}). \\ \hline

18 & \texttt{GS} & $e_{RE}$ (\ref{eq:angle_rotmat}) & 
Conversion to rotation matrix using \eqref{eq:fGS}.
\\ \hline

19 & \texttt{GS} & $L2 + L2$ (\ref{eqn:l2}) & 
Loss is the sum of $L2$ losses for the two vectors.  \\ \hline

20 & \texttt{GS} & $e_{TE} + e_{TE}$ (\ref{eq:vector_angle}) &
Loss is the sum of $e_{TE}$ losses for the two vectors. \\ \hline

\hline
    \end{tabular}
    }
    \caption{The table lists all of the evaluated models with different representations and loss functions including implementation details. For some conversions we use the kornia library~\cite{kornia}.}
    \label{tab:implement}
\end{table}

%% file: chapters/50-results.tex
\vspace{-5mm}
\section{Experiments and Evaluation}
\label{section:results}
In this section, we compare the results for the different representations and loss functions. Each model was trained for $100$ epochs using the Adam optimizer~\cite{adam}. 

\subsection{Error Metric}
\label{def:mAA}
We use the rotation angle $e_{RE}$ between the actual and predicted matrices as defined in \eqref{eq:angle_rotmat} to evaluate the network performance. The results will be presented in two ways. First, on graphs in the form of \textbf{accuracy curves}, where the $x$-axis is the magnitude of the error angle, and the $y$-axis is the fraction of test data points that had a test error angle smaller than a given threshold on the $x$-axis.

Based on this plot, we compute the mean average accuracy (mAA) \cite{auc} with a cutoff value of $\alpha$. mAA is the normalized area under the curve on the interval $(0, \alpha)$. The higher the value of mAA, the larger the proportion of smaller errors, and hence on average the more accurate the method. 

\subsection{Synthetic Data Results}

We will start by presenting results on our generated data sets (described in Section \ref{section:generation}). This will be followed by results on real data in the next subsection.

\subsubsection{Impact of different distributions}

\begin{figure}[h]
    \centering
        \includegraphics[width=1.0\linewidth]{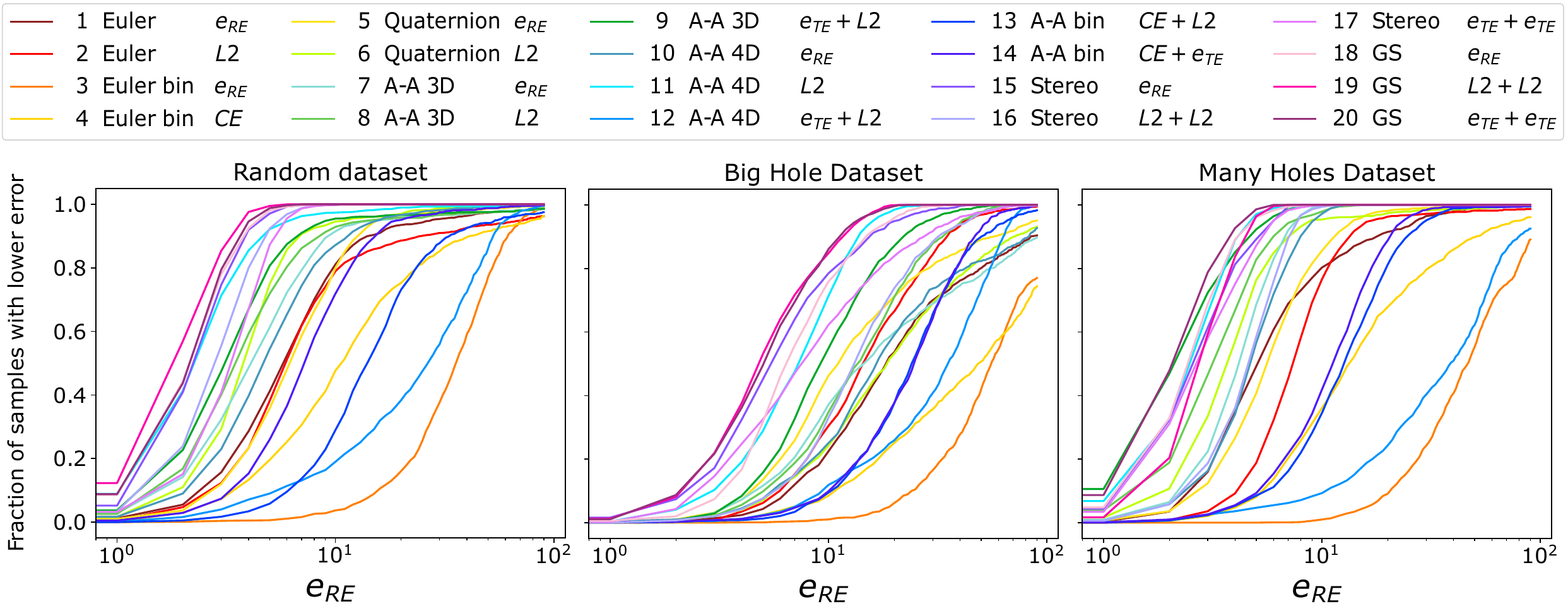}
    \caption{Accuracy curves for rotation error distributions within different datasets.}    
    \label{fig:roc_distr}
\end{figure}

We have trained the network on different different data distributions, with the design of training and test splits described in Section~\ref{section:generation}.

In Fig.~\ref{fig:roc_distr} we present the accuracy curves for each method. For better clarity, we present this plot with a logarithmic $x$-axis. We also report the results of the methods for each distribution in tables \ref{tab:random}, \ref{tab:many_holes} and \ref{tab:big_hole}. The best values for each column are highlighted in the table.  For all distributions, the dominance of higher dimensional representations and, conversely, the poor performance of the classification models can be seen. 

The difference in results between the random distribution and the one with many holes is not that big. We can see in the graph that there are methods that are more accurate on this dataset (higher density of curves with steeper slope at the beginning of the graph) but also methods that deliver worse results.  Of interest is the success of the $3$ dimensional axis-angle representation (Id=$9$).

However, when compared to the large hole dataset, we see a more pronounced difference. 
As reflected in the higher test errors, the networks had difficulty generalizing to this unknown region of the data distribution. 



\begin{table}[htbp]
\centering
\resizebox{\textwidth}{!}{
\begin{tabular}{|c c c c c c c c|}
\hline
Id & Representation & Loss function & Mean & Median & mAA($5^\circ$) & mAA($10^\circ$) & mAA($20^\circ$) \\
\hline
1 & \texttt{Euler} & $e_{RE}$ (\ref{eq:angle_rotmat}) & 9.82 & 5.72 & 0.18 & 0.43 & 0.67 \\
2 & \texttt{Euler} & $L2$ (\ref{eqn:l2}) & 13.71 & 5.93 & 0.16 & 0.41 & 0.64 \\
3 & \texttt{Euler bin} & $e_{RE}$ (\ref{eq:angle_rotmat}) & 38.12 & 35.61 & 0.00 & 0.01 & 0.06 \\
4 & \texttt{Euler bin} & $CE$ (\ref{eq:CE}) & 19.70 & 10.64 & 0.09 & 0.22 & 0.44 \\
5 & \texttt{Quaternion} & $e_{RE}$ (\ref{eq:angle_rotmat}) & 8.42 & 6.15 & 0.15 & 0.40 & 0.67 \\
6 & \texttt{Quaternion} & $L2$ (\ref{eqn:l2}) & 5.27 & 3.84 & 0.34 & 0.63 & 0.79 \\
7 & \texttt{A-A 3D} & $e_{RE}$ (\ref{eq:angle_rotmat}) & 7.73 & 4.09 & 0.32 & 0.58 & 0.76 \\
8 & \texttt{A-A 3D} & $L2$ (\ref{eqn:l2}) & 6.87 & 3.54 & 0.38 & 0.63 & 0.79 \\
9 & \texttt{A-A 3D} & $e_{TE}$ (\ref{eq:vector_angle}) + $L2$ (\ref{eqn:l2}) & 6.00 & 3.12 & 0.44 & 0.68 & 0.82 \\
10 & \texttt{A-A 4D} & $e_{RE}$ (\ref{eq:angle_rotmat}) & 6.82 & 4.68 & 0.26 & 0.52 & 0.73 \\
11 & \texttt{A-A 4D} & $L2$ (\ref{eqn:l2}) & 3.46 & \a{2.25} & \a{0.60} & \a{0.78} & \a{0.88} \\
12 & \texttt{A-A 4D} & $e_{TE}$ (\ref{eq:vector_angle}) + $L2$ (\ref{eqn:l2}) & 28.51 & 26.60 & 0.04 & 0.10 & 0.20 \\
13 & \texttt{A-A bin} & $CE$ (\ref{eq:CE}) + $L2$ (\ref{eqn:l2}) & 19.48 & 13.98 & 0.02 & 0.10 & 0.33 \\
14 & \texttt{A-A bin} & $CE$ (\ref{eq:CE}) + $e_{TE}$ (\ref{eq:vector_angle}) & 9.30 & 7.09 & 0.10 & 0.33 & 0.61 \\
15 & \texttt{Stereo} & $e_{RE}$ (\ref{eq:angle_rotmat}) & \a{2.41} & \a{2.21} & \a{0.62} & \a{0.81} & \a{0.90} \\
16 & \texttt{Stereo} & $L2 + L2$ (\ref{eqn:l2}) & \a{3.01} & 2.81 & 0.51 & 0.75 & \a{0.88} \\
17 & \texttt{Stereo} & $e_{TE} + e_{TE}$ (\ref{eq:vector_angle}) & 3.45 & 3.32 & 0.43 & 0.71 & 0.85 \\
18 & \texttt{GS} & $e_{RE}$ (\ref{eq:angle_rotmat}) & \a{2.29} & \a{2.13} & \a{0.64} & \a{0.82} & \a{0.91} \\
19 & \texttt{GS} & $L2 + L2$ (\ref{eqn:l2}) & \b{1.99} & \b{1.87} & \b{0.70} & \b{0.85} &\b{ 0.93} \\
20 & \texttt{GS} & $e_{TE} + e_{TE}$ (\ref{eq:vector_angle}) & \a{2.26} & \a{2.13} & \a{0.65} & \a{0.83} & \a{0.91} \\
\hline
\end{tabular}
}
\caption{Test set results for the dataset with random data distribution.}
\label{tab:random}
\end{table}


\begin{table}[htbp]
\centering
\resizebox{\textwidth}{!}{
\begin{tabular}{|c c c c c c c c|}
\hline
Id & Representation & Loss function & Mean & Median & mAA($5^\circ$) & mAA($10^\circ$) & mAA($20^\circ$) \\
\hline
1 & \texttt{Euler} & $e_{RE}$ (\ref{eq:angle_rotmat}) & 8.27 & 5.02 & 0.21 & 0.46 & 0.67 \\
2 & \texttt{Euler} & $L2$ (\ref{eqn:l2}) & 10.20 & 7.38 & 0.06 & 0.31 & 0.62 \\
3 & \texttt{Euler bin} & $e_{RE}$ (\ref{eq:angle_rotmat}) & 51.56 & 46.53 & 0.00 & 0.00 & 0.03 \\
4 & \texttt{Euler bin} & $CE$ (\ref{eq:CE}) & 22.93 & 13.32 & 0.03 & 0.14 & 0.35 \\
5 & \texttt{Quaternion} & $e_{RE}$ (\ref{eq:angle_rotmat}) & 7.03 & 5.61 & 0.17 & 0.45 & 0.70 \\
6 & \texttt{Quaternion} & $L2$ (\ref{eqn:l2}) & 5.39 & 3.71 & 0.36 & 0.64 & 0.80 \\
7 & \texttt{A-A 3D} & $e_{RE}$ (\ref{eq:angle_rotmat}) & 4.36 & 4.13 & 0.29 & 0.61 & 0.81 \\
8 & \texttt{A-A 3D} & $L2$ (\ref{eqn:l2}) & 3.59 & 3.15 & 0.44 & 0.69 & 0.85 \\
9 & \texttt{A-A 3D} & $e_{TE}$ (\ref{eq:vector_angle}) + $L2$ (\ref{eqn:l2}) & \a{2.49} & \a{2.08} & \a{0.61} & \a{0.80} & \a{0.90} \\
10 & \texttt{A-A 4D} & $e_{RE}$ (\ref{eq:angle_rotmat}) & 5.02 & 4.72 & 0.23 & 0.55 & 0.77 \\
11 & \texttt{A-A 4D} & $L2$ (\ref{eqn:l2}) & \a{2.61} & \a{2.44} & \a{0.58} & \a{0.79} & \a{0.89} \\
12 & \texttt{A-A 4D} & $e_{TE}$ (\ref{eq:vector_angle}) + $L2$ (\ref{eqn:l2}) & 42.12 & 38.60 & 0.02 & 0.05 & 0.11 \\
13 & \texttt{A-A bin} & $CE$ (\ref{eq:CE}) + $L2$ (\ref{eqn:l2}) & 14.48 & 12.52 & 0.03 & 0.13 & 0.38 \\
14 & \texttt{A-A bin} & $CE$ (\ref{eq:CE}) + $e_{TE}$ (\ref{eq:vector_angle}) & 12.51 & 11.18 & 0.04 & 0.16 & 0.44 \\
15 & \texttt{Stereo} & $e_{RE}$ (\ref{eq:angle_rotmat}) & \a{2.86} & \a{2.55} & \a{0.54} & \a{0.76} & \a{0.88} \\
16 & \texttt{Stereo} & $L2 + L2$ (\ref{eqn:l2}) & 4.70 & 4.60 & 0.24 & 0.58 & 0.79 \\
17 & \texttt{Stereo} & $e_{TE} + e_{TE}$ (\ref{eq:vector_angle}) & 3.03 & 2.62 & 0.51 & 0.75 & 0.87 \\
18 & \texttt{GS} & $e_{RE}$ (\ref{eq:angle_rotmat}) & \a{2.60} & \a{2.45} & \a{0.58} & \a{0.79} & \a{0.89} \\
19 & \texttt{GS} & $L2 + L2$ (\ref{eqn:l2}) & 2.87 & 2.71 & 0.52 & \a{0.76} & 0.88 \\
20 & \texttt{GS} & $e_{TE} + e_{TE}$ (\ref{eq:vector_angle}) & \b{2.24} & \b{2.05} & \b{0.65} & \b{0.83} & \b{0.91} \\
\hline
\end{tabular}
}
\caption{Test set results for the dataset with many holes.}
\label{tab:many_holes}
\end{table}


\begin{table}[htbp]
\centering
\resizebox{\textwidth}{!}{
\begin{tabular}{|c c c c c c c c|}
\hline
Id & Representation & Loss function & Mean & Median & mAA($5^\circ$) & mAA($10^\circ$) & mAA($20^\circ$) \\
\hline
1 & \texttt{Euler} & $e_{RE}$ (\ref{eq:angle_rotmat}) & 33.30 & 18.23 & 0.02 & 0.08 & 0.25 \\
2 & \texttt{Euler} & $L2$ (\ref{eqn:l2}) & 18.45 & 14.30 & 0.02 & 0.11 & 0.32 \\
3 & \texttt{Euler bin} & $e_{RE}$ (\ref{eq:angle_rotmat}) & 65.44 & 55.50 & 0.00 & 0.00 & 0.02 \\
4 & \texttt{Euler bin} & $CE$ (\ref{eq:CE}) & 61.58 & 48.18 & 0.01 & 0.03 & 0.11 \\
5 & \texttt{Quaternion} & $e_{RE}$ (\ref{eq:angle_rotmat}) & 22.43 & 11.01 & 0.05 & 0.19 & 0.41 \\
6 & \texttt{Quaternion} & $L2$ (\ref{eqn:l2}) & 30.63 & 18.47 & 0.03 & 0.10 & 0.26 \\
7 & \texttt{A-A 3D} & $e_{RE}$ (\ref{eq:angle_rotmat}) & 32.01 & 14.44 & 0.04 & 0.15 & 0.34 \\
8 & \texttt{A-A 3D} & $L2$ (\ref{eqn:l2}) & 16.68 & 13.52 & 0.04 & 0.14 & 0.35 \\
9 & \texttt{A-A 3D} & $e_{TE}$ (\ref{eq:vector_angle}) + $L2$ (\ref{eqn:l2}) & 11.69 & 9.15 & 0.06 & 0.23 & 0.50 \\
10 & \texttt{A-A 4D} & $e_{RE}$ (\ref{eq:angle_rotmat}) & 29.39 & 16.51 & 0.03 & 0.11 & 0.28 \\
11 & \texttt{A-A 4D} & $L2$ (\ref{eqn:l2}) & \a{7.94} & \a{7.30} & 0.13 & \a{0.34} & \a{0.63} \\
12 & \texttt{A-A 4D} & $e_{TE}$ (\ref{eq:vector_angle}) + $L2$ (\ref{eqn:l2}) & 35.96 & 35.20 & 0.01 & 0.04 & 0.12 \\
13 & \texttt{A-A bin} & $CE$ (\ref{eq:CE}) + $L2$ (\ref{eqn:l2}) & 28.16 & 24.01 & 0.01 & 0.03 & 0.14 \\
14 & \texttt{A-A bin} & $CE$ (\ref{eq:CE}) + $e_{TE}$ (\ref{eq:vector_angle}) & 26.82 & 24.63 & 0.01 & 0.04 & 0.14 \\
15 & \texttt{Stereo} & $e_{RE}$ (\ref{eq:angle_rotmat}) & \a{7.49} & \a{5.43} & \a{0.21} & \a{0.45} & \a{0.67} \\
16 & \texttt{Stereo} & $L2 + L2$ (\ref{eqn:l2}) & 16.67 & 13.23 & 0.03 & 0.12 & 0.35 \\
17 & \texttt{Stereo} & $e_{TE} + e_{TE}$ (\ref{eq:vector_angle}) & 11.87 & 7.52 & \a{0.15} & \a{0.34} & 0.55 \\
18 & \texttt{GS} & $e_{RE}$ (\ref{eq:angle_rotmat}) & \a{8.00} & \a{6.43} & \a{0.12} & \a{0.37} & \a{0.63} \\
19 & \texttt{GS} & $L2 + L2$ (\ref{eqn:l2}) & \b{5.90} & \b{4.81} & \b{0.24} & \b{0.50} & \b{0.73} \\
20 & \texttt{GS} & $e_{TE} + e_{TE}$ (\ref{eq:vector_angle}) & \a{5.97} & \a{5.04} & \a{0.23} & \a{0.49} & \b{0.73} \\
\hline
\end{tabular}
}
\caption{Test set results for the dataset with single big hole. \vspace{-5mm}}
\label{tab:big_hole}
\end{table}



\begin{figure}
    \centering
    \includegraphics[height=40mm]{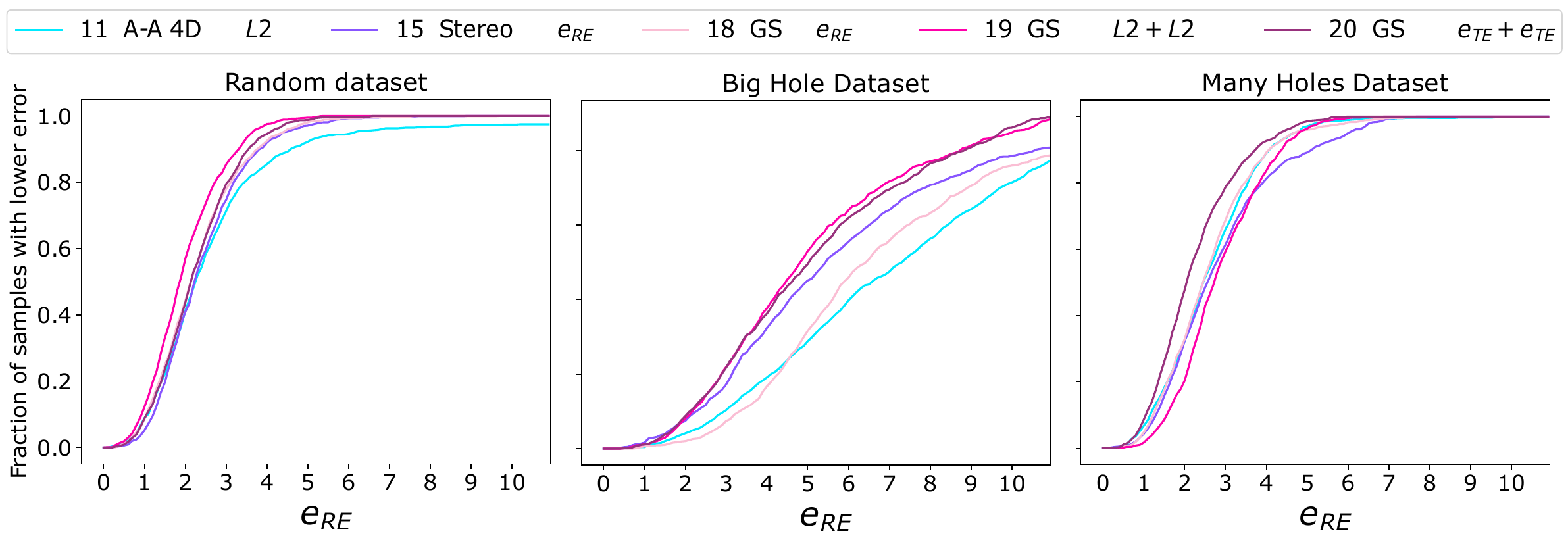}
    \caption{Acc. curves for the 5 best methods for each distribution.}
    \label{fig:roc_distr:top5}
\end{figure}

In Fig. \ref{fig:roc_distr:top5} we show the accuracy curve within the $10^\circ$ bound for the selection of the $5$ best methods. 
Interestingly, there is a good agreement between the relative rankings within the random distribution and the large-hole distribution, whereas the performance differs slightly for the dataset with many holes. 
This suggests that there may not be an universally best method. 

Although we have only $3$ datasets with different difficulty values $\zeta$ \eqref{def:zeta}, we see a clear dependence of the network results on $\zeta$ for all methods.

\begin{figure}[H]
    \centering
    \vspace{-3mm}\includegraphics[width=0.9\linewidth]{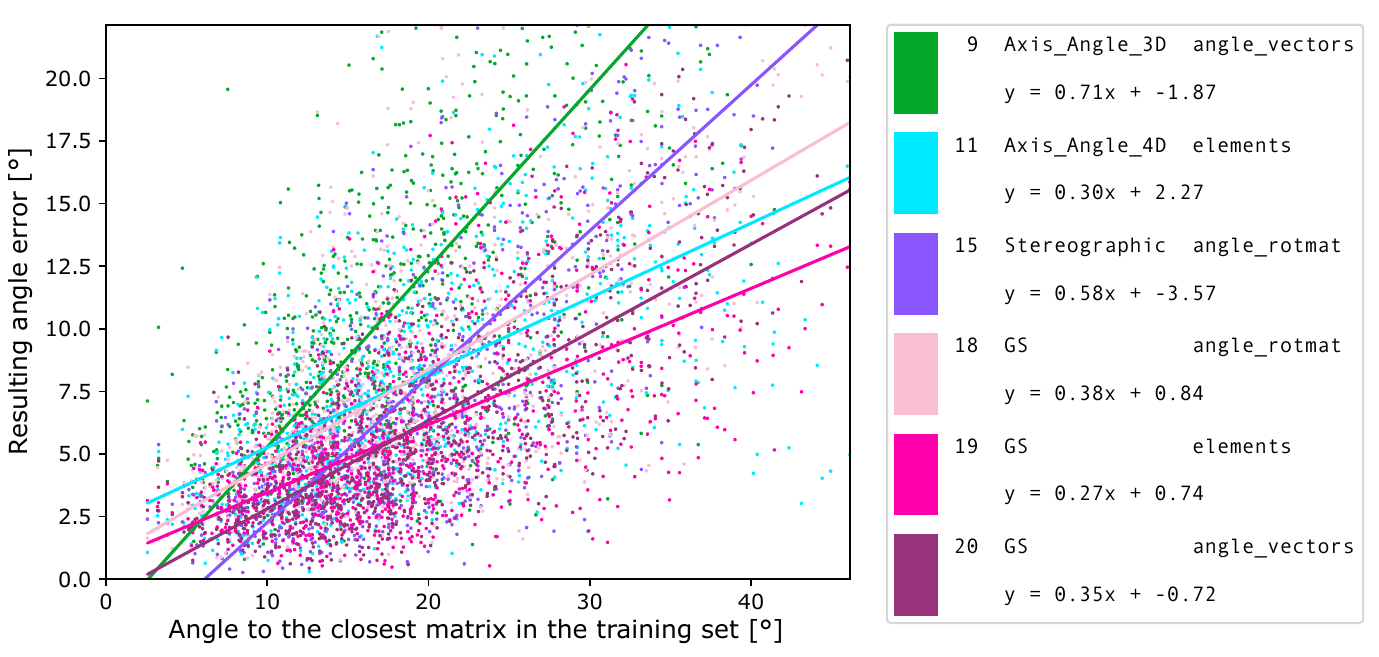}
    \caption{Error angle for each in respect to the distance from the training set for each matrix in the test set and each of the $6$ best methods, evaluated on the Big Hole Dataset.}
    \label{fig:closest_vs_err}
\end{figure}

In Fig. \ref{fig:closest_vs_err}, we show $e_{RE}$ for each matrix in the test set as a function of the minimum distance of that matrix from the training set for the large hole dataset. We plot these data points for each of the $6$ best methods and then perform linear fit with the least squares method to evaluate their relationship. 
In this case, straight lines with a steeper slope imply worse robustness of the network. This shows that the 6D GS representation with (Id=19) and the axis-angle 4D (Id=11) representation exhibit the best robustness.



\subsubsection{Impact of various object textures}
\begin{figure}
    \centering

    \includegraphics[width=1.0\linewidth]{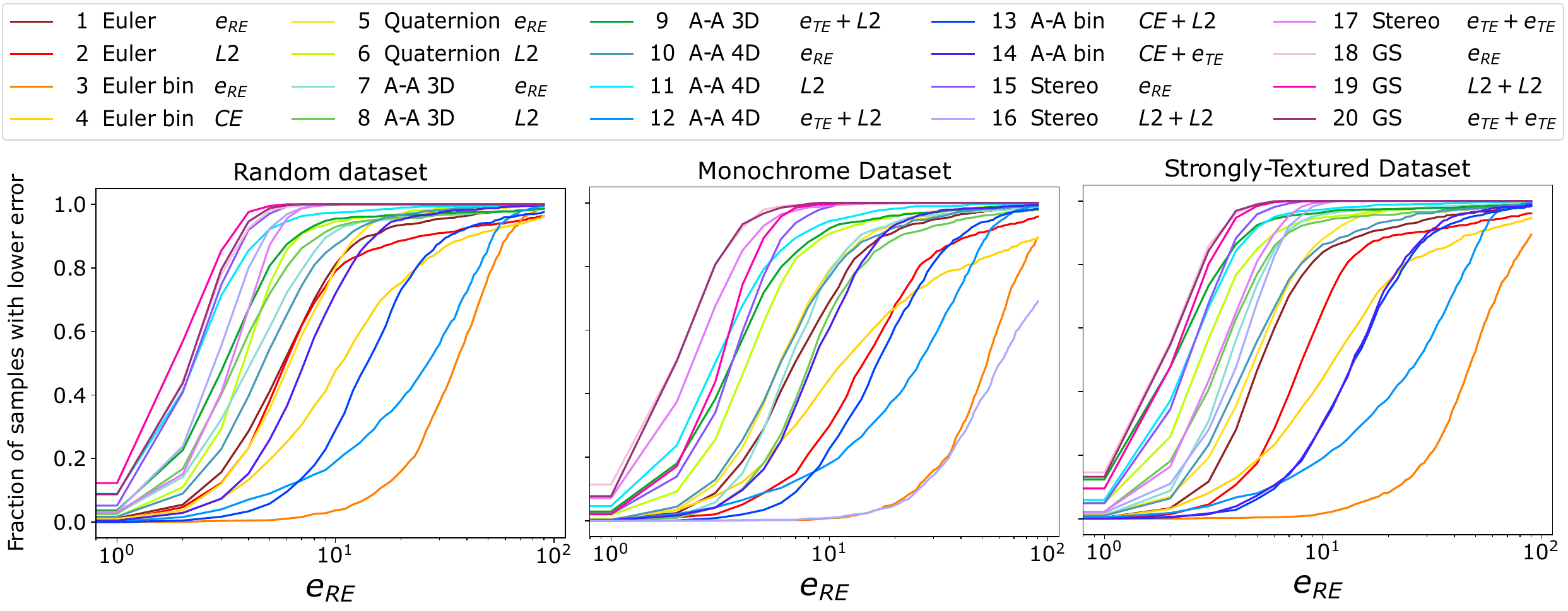}
    \vspace{-5mm}
    \caption{Accuracy curves for individual texture-styles in log scale.}
    \label{fig:roc_textures}
\end{figure}

Next, we investigate the effect of texture on the performance of the networks. 
We report the results for each texture style in tables \ref{tab:random}, \ref{tab:textured} and \ref{tab:monochrome}. The best values for each column are highlighted in the table.  Again, we also show Figure \ref{fig:roc_textures} with the accuracy curves for each method and texture.
We see that even across this set of experiments, the representations with higher number of variables dominate. When comparing the random and strongly-textured datasets, we do not see large differences.

For the plots of the monochrome dataset, larger changes are visible. Although the 6D and 5D representations more or less retain good results, several methods have a less steep slope and thus a worse robustness. The only significantly different curve is that of the 5D method (Id=16). Overall, only the monochromatic dataset has significantly different results.

\begin{table}[htbp]
\centering
\resizebox{11.5cm}{!}{
\begin{tabular}{|c c c c c c c c|}
\hline
Id & Representation & Loss function & Mean & Median & mAA($5^\circ$) & mAA($10^\circ$) & mAA($20^\circ$) \\
\hline
1 & \texttt{Euler} & $e_{RE}$ (\ref{eq:angle_rotmat}) & 10.16 & 5.23 & 0.18 & 0.46 & 0.68 \\
2 & \texttt{Euler} & $L2$ (\ref{eqn:l2}) & 15.64 & 8.04 & 0.07 & 0.28 & 0.56 \\
3 & \texttt{Euler bin} & $e_{RE}$ (\ref{eq:angle_rotmat}) & 53.47 & 48.12 & 0.00 & 0.01 & 0.03 \\
4 & \texttt{Euler bin} & $CE$ (\ref{eq:CE}) & 22.35 & 11.29 & 0.09 & 0.21 & 0.43 \\
5 & \texttt{Quaternion} & $e_{RE}$ (\ref{eq:angle_rotmat}) & 7.00 & 4.88 & 0.23 & 0.51 & 0.73 \\
6 & \texttt{Quaternion} & $L2$ (\ref{eqn:l2}) & 5.01 & 2.77 & 0.50 & 0.71 & 0.84 \\
7 & \texttt{A-A 3D} & $e_{RE}$ (\ref{eq:angle_rotmat}) & 5.74 & 3.74 & 0.34 & 0.63 & 0.80 \\
8 & \texttt{A-A 3D} & $L2$ (\ref{eqn:l2}) & 6.68 & 3.35 & 0.40 & 0.66 & 0.80 \\
9 & \texttt{A-A 3D} & $e_{TE}$ (\ref{eq:vector_angle}) + $L2$ (\ref{eqn:l2}) & 4.52 & \a{2.09} & \a{0.62} & \a{0.79} & \a{0.88} \\
10 & \texttt{A-A 4D} & $e_{RE}$ (\ref{eq:angle_rotmat}) & 8.26 & 4.56 & 0.26 & 0.52 & 0.71 \\
11 & \texttt{A-A 4D} & $L2$ (\ref{eqn:l2}) & \a{3.40} & 2.42 & 0.57 & 0.77 & \a{0.88} \\
12 & \texttt{A-A 4D} & $e_{TE}$ (\ref{eq:vector_angle}) + $L2$ (\ref{eqn:l2}) & 28.96 & 27.21 & 0.04 & 0.10 & 0.20 \\
13 & \texttt{A-A bin} & $CE$ (\ref{eq:CE}) + $L2$ (\ref{eqn:l2}) & 18.74 & 13.98 & 0.02 & 0.11 & 0.33 \\
14 & \texttt{A-A bin} & $CE$ (\ref{eq:CE}) + $e_{TE}$ (\ref{eq:vector_angle}) & 17.72 & 13.85 & 0.03 & 0.11 & 0.34 \\
15 & \texttt{Stereo} & $e_{RE}$ (\ref{eq:angle_rotmat}) & \a{2.61} & \a{2.43} & \a{0.58} & \a{0.79} & \a{0.89} \\
16 & \texttt{Stereo} & $L2 + L2$ (\ref{eqn:l2}) & 4.25 & 4.09 & 0.31 & 0.62 & 0.81 \\
17 & \texttt{Stereo} & $e_{TE} + e_{TE}$ (\ref{eq:vector_angle}) & 3.60 & 3.32 & 0.42 & 0.69 & 0.85 \\
18 & \texttt{GS} & $e_{RE}$ (\ref{eq:angle_rotmat}) & \b{1.98} & \b{1.86} & \b{0.70} & \b{0.85} & \b{0.93} \\
19 & \texttt{GS} & $L2 + L2$ (\ref{eqn:l2}) & \a{2.21} & \a{2.05} & \a{0.66} & \a{0.83} & \a{0.91} \\
20 & \texttt{GS} & $e_{TE} + e_{TE}$ (\ref{eq:vector_angle}) & \a{2.03} & \a{1.90} & \b{0.70} & \b{0.85} & \a{0.92} \\
\hline
\end{tabular}
}
\caption{Test set results for the strongly-textured dataset.}
\label{tab:textured}
\end{table}


\begin{table}[htbp]
\centering
\resizebox{11.5cm}{!}{
\begin{tabular}{|c c c c c c c c|}
\hline
Id & Representation & Loss function & Mean & Median & mAA($5^\circ$) & mAA($10^\circ$) & mAA($20^\circ$) \\
\hline
1 & \texttt{Euler} & $e_{RE}$ (\ref{eq:angle_rotmat}) & 10.95 & 6.98 & 0.12 & 0.34 & 0.60 \\
2 & \texttt{Euler} & $L2$ (\ref{eqn:l2}) & 22.84 & 14.48 & 0.03 & 0.12 & 0.32 \\
3 & \texttt{Euler bin} & $e_{RE}$ (\ref{eq:angle_rotmat}) & 56.54 & 52.78 & 0.00 & 0.00 & 0.02 \\
4 & \texttt{Euler bin} & $CE$ (\ref{eq:CE}) & 30.44 & 11.61 & 0.08 & 0.22 & 0.40 \\
5 & \texttt{Quaternion} & $e_{RE}$ (\ref{eq:angle_rotmat}) & 9.39 & 5.96 & 0.15 & 0.41 & 0.65 \\
6 & \texttt{Quaternion} & $L2$ (\ref{eqn:l2}) & 6.80 & 4.22 & 0.29 & 0.57 & 0.75 \\
7 & \texttt{A-A 3D} & $e_{RE}$ (\ref{eq:angle_rotmat}) & 10.32 & 6.53 & 0.10 & 0.36 & 0.63 \\
8 & \texttt{A-A 3D} & $L2$ (\ref{eqn:l2}) & 13.79 & 8.18 & 0.07 & 0.27 & 0.54 \\
9 & \texttt{A-A 3D} & $e_{TE}$ (\ref{eq:vector_angle}) + $L2$ (\ref{eqn:l2}) & 6.35 & 3.57 & 0.38 & 0.62 & 0.78 \\
10 & \texttt{A-A 4D} & $e_{RE}$ (\ref{eq:angle_rotmat}) & 9.72 & 6.00 & 0.16 & 0.41 & 0.64 \\
11 & \texttt{A-A 4D} & $L2$ (\ref{eqn:l2}) & 4.54 & 3.02 & 0.45 & 0.68 & 0.82 \\
12 & \texttt{A-A 4D} & $e_{TE}$ (\ref{eq:vector_angle}) + $L2$ (\ref{eqn:l2}) & 28.54 & 25.89 & 0.04 & 0.09 & 0.20 \\
13 & \texttt{A-A bin} & $CE$ (\ref{eq:CE}) + $L2$ (\ref{eqn:l2}) & 22.06 & 16.59 & 0.01 & 0.07 & 0.26 \\
14 & \texttt{A-A bin} & $CE$ (\ref{eq:CE}) + $e_{TE}$ (\ref{eq:vector_angle}) & 11.23 & 8.45 & 0.06 & 0.25 & 0.54 \\
15 & \texttt{Stereo} & $e_{RE}$ (\ref{eq:angle_rotmat}) & \a{3.90} & \a{3.64} & \a{0.37} & \a{0.66} & \a{0.83} \\
16 & \texttt{Stereo} & $L2 + L2$ (\ref{eqn:l2}) & 90.45 & 87.58 & 0.00 & 0.00 & 0.01 \\
17 & \texttt{Stereo} & $e_{TE} + e_{TE}$ (\ref{eq:vector_angle}) & \a{2.72} & \a{2.35} & \a{0.58} & \a{0.78} & \a{0.89} \\
18 & \texttt{GS} & $e_{RE}$ (\ref{eq:angle_rotmat}) & \b{2.23} & \b{1.99} & \b{0.67} & \b{0.83} & \b{0.91} \\
19 & \texttt{GS} & $L2 + L2$ (\ref{eqn:l2}) & \a{3.34} & \a{3.21} & \a{0.45} & \a{0.72} & \a{0.86} \\
20 & \texttt{GS} & $e_{TE} + e_{TE}$ (\ref{eq:vector_angle}) & \b{2.23} & \b{1.99} & \a{0.66} & \b{0.83} & \b{0.91} \\
\hline
\end{tabular}
}
\caption{Test set results for dataset with monochromatic texture.}
\label{tab:monochrome}
\end{table}

\subsubsection{Synthetic Data Results Summary}
In all cases, we see consistently better results for higher-dimensional representations, in particular \texttt{GS} and \texttt{Stereo}. However, some implementations using the Axis-angle representation also stood out. On the contrary, the classification model performed poorly. 

\begin{figure}
    \centering
    \includegraphics[width=0.8\linewidth]{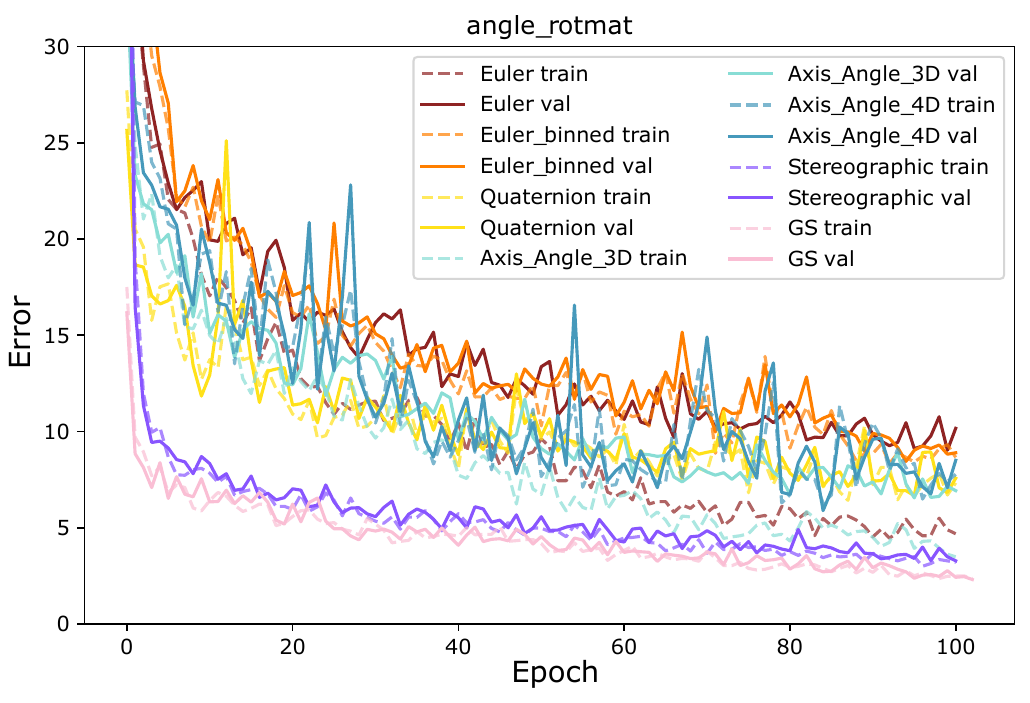}
    \caption{Evolution of training and validation error during training on the strongly-textured dataset for each representation.}
    \label{fig:angle_rotmat}
\end{figure}

The results do not point to a single best-performing loss function. However, when training neural networks the speed of convergence is also important. Fig.~\ref{fig:angle_rotmat} shows the evolution of mean $e_{RE}$ on the training and validation sets for the strongly colored dataset with the random distribution.
We can see that also in this case, the 6D GS and 5D Stereo representations achieve the lowest training and validation error. Moreover, the low values were acquired very quickly after the start of training, in contrast to the other representations. 


%% file: chapters/60-boxes.tex
\subsection{Real Data Results}
\begin{figure}[h]
    \centering
    \vspace{-8mm}
    \includegraphics[width=0.8\linewidth]{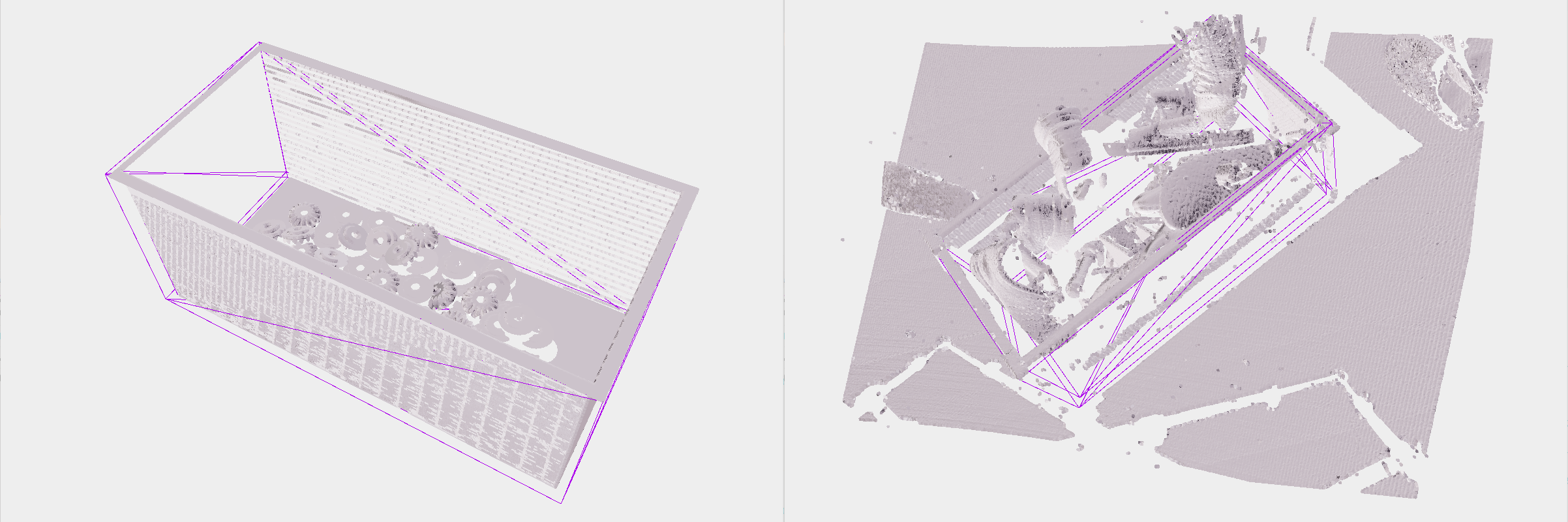}
    \vspace{-1mm}
    \caption{Examples from the dataset with bins, ground-truth bin positions are displayed in purple.}
    \label{fig:pointers_bins}
\end{figure}

In this section, we apply our implementation to the real-world problem of determining the pose of a bin based on its 3D scan. We use the dataset provided in \cite{prepravky} for evaluation. This dataset consists of two kinds of data: scans of real bins and  synthetically generated data. 
The input is a structured point cloud in the form of XYZ coordinates, which are assigned gray-scale intensity texture. Additionally, these scans were annotated using a semi-automated method that in some cases required users to mark the corners of the bins in a special application. Examples from this dataset with ground truth positions of bins can be seen in Fig. \ref{fig:pointers_bins}. On the left we see the synthetically generated scan and on the right a real scan.

We adapted our network to this data. We evaluated the $5$ methods which performed best on our synthetic examples as presented in the previous section. During training we follow the procedure in \cite{prepravky} to deal with the bin symmetry. 

We performed 2 experiments. In one we used only data that was not synthetic. 
There were 384 scans in the training set and 39 in the test set. In the second experiment, we used all available data. Here the amount of samples in the training set was 672 and in the test set 49.

We present the results in Tables \ref{tab:bins_real} and \ref{tab:bins_synth}. Unlike previous results, here we present the result of the trained model from the epoch that had the smallest validation error. We ran $200$ epochs for each method in this experiment.


\begin{table}
    \centering
     \resizebox{11.0cm}{!}{
\begin{tabular}{|c c c c c c c c|}
\hline
Id & Representation & Loss function & Mean & Median & mAA($5^\circ$) & mAA($10^\circ$) & mAA($20^\circ$) \\
    \hline
    11 & \texttt{A-A 4D} & $L2$ (\ref{eqn:l2}) & 57.94 & 42.68 & 0.01 & 0.02 & 0.08 \\
    15 & \texttt{Stereo} & $e_{RE}$ (\ref{eq:angle_rotmat}) & \a{23.35} & \a{15.22} & \b{0.08} & \a{0.16} & \a{0.34} \\
    18 & \texttt{GS} & $e_{RE}$ (\ref{eq:angle_rotmat}) & \b{23.16} & \b{13.93} & \a{0.05} & \b{0.17} & \b{0.38} \\
    19 & \texttt{GS} & $L2 + L2$ (\ref{eqn:l2}) & 27.97 & 23.38 & 0.02 & 0.07 & 0.17 \\
    20 & \texttt{GS} & $e_{TE} + e_{TE}$ (\ref{eq:vector_angle}) & \a{25.26} & \a{15.32} & \a{0.03} & \a{0.14} & \a{0.31} \\
    \hline
\end{tabular}
}
    \caption{Results for selected methods on the bin pose estimation dataset~\cite{prepravky} using only non-synthetic data. \vspace{-10mm}}
    \label{tab:bins_real}
\end{table}

\begin{table}
    \centering
     \resizebox{11.0cm}{!}{
\begin{tabular}{|c c c c c c c c|}
\hline
Id & Representation & Loss function & Mean & Median & mAA($5^\circ$) & mAA($10^\circ$) & mAA($20^\circ$) \\
    \hline
    11 & \texttt{A-A 4D} & $L2$ (\ref{eqn:l2}) & 41.99 & 27.46 & 0.02 & 0.07 & 0.18 \\
    15 & \texttt{Stereo} & $e_{RE}$ (\ref{eq:angle_rotmat}) & \a{18.64} & \a{9.54} & \b{0.11} & \b{0.27} & \a{0.45} \\
    18 & \texttt{GS} & $e_{RE}$ (\ref{eq:angle_rotmat}) & \b{18.41} & \b{8.32} & \a{0.07} & \a{0.26} & \b{0.48} \\
    19 & \texttt{GS} & $L2 + L2$ (\ref{eqn:l2}) & 21.65 & 16.32 & 0.09 & 0.19 & 0.35 \\
    20 & \texttt{GS} & $e_{TE} + e_{TE}$ (\ref{eq:vector_angle}) & \a{19.51} & \a{10.44} & \a{0.10} & \a{0.26} & \a{0.45} \\
    \hline
\end{tabular}
}
    \caption{Results for selected methods on the bin pose estimation dataset~\cite{prepravky} using all available data.}
    \label{tab:bins_synth}
\end{table}  

The results show that the axis angle 4D representation performs worse than GS 6D and stereographic 5D. The $e_{RE}$ loss function shows slightly better results compared to other used loss functions.


%% file: chapters/90-conclusion.tex
\section{Conclusion} 

In this paper, we presented several synthetic and real experiments aimed at evaluating different potential representations and loss functions for 3D rotation estimation. In line with previous research~\cite{rot3d}, the methods that used higher dimensional representations, namely the 6D Gram-Schmidt and 5D streographic representations performed the best. Among the discontinuous representations, the method using the 4D representation of the Axis-angle achieved the best results. The 6D Gram-Schmidt representation performs well with all of the evaluated loss functions, while 5D stereographic representation performs best with $e_{RE}$ loss and the Axis-angle representation performs best with the $L2$ loss.

The networks were evaluated only for one training run for each combination of dataset, representation and loss function. Since training neural networks includes a degree of randomness process, it would be beneficial in the future to perform multiple training runs for each method and present some summary statistics. Furthermore, the fact that we restricted our experiments to one kind of object may bias the results. For the results on the real dataset, we did not address the various translations of the object within the scene. Despite these limitations, the results presented in this paper shed a light on the effects of different rotation representations and loss functions on 3D rotation estimation.
\vspace{-5mm}

%% file: authorsample.bbl
\begin{thebibliography}{10}

\bibitem{Alzubaidi2021}
Laith Alzubaidi, Jinglan Zhang, Amjad~J. Humaidi, Ayad Al-dujaili, Ye~Duan, Omran Al-Shamma, Jos{\'e}~I. Santamar{\'i}a, Mohammed~Abdulraheem Fadhel, Muthana Al-Amidie, and Laith Farhan.
\newblock Review of deep learning: concepts, cnn architectures, challenges, applications, future directions.
\newblock {\em Journal of Big Data}, 8, 2021.

\bibitem{Chen2021}
Wei Chen, Xi~Jia, Hyung~Jin Chang, Jinming Duan, Linlin Shen, and Ales Leonardis.
\newblock Fs-net: Fast shape-based network for category-level 6d object pose estimation with decoupled rotation mechanism.
\newblock In {\em Proceedings of the IEEE/CVF Conference on Computer Vision and Pattern Recognition}, pages 1581--1590, 2021.

\bibitem{prepravky}
Luk{\'a}{\v{s}} Gajdo{\v{s}}ech, Viktor Kocur, Martin Stuchl\'{i}k, Luk{\'a}{\v{s}} Hudec, and Martin Madaras.
\newblock Towards deep learning-based 6d bin pose estimation in 3d scans.
\newblock In {\em Proceedings of the 17th International Joint Conference on Computer Vision, Imaging and Computer Graphics Theory and Applications}, pages 545--552. Scitepress, February 2022.

\bibitem{fibonaci_spiral}
Álvaro González.
\newblock Measurement of areas on a sphere using fibonacci and latitude–longitude lattices.
\newblock {\em Mathematical Geosciences}, 42(1):49–64, November 2009.

\bibitem{Hartley2012}
Richard~I. Hartley, Jochen Trumpf, Yuchao Dai, and Hongdong Li.
\newblock Rotation averaging.
\newblock {\em International Journal of Computer Vision}, 103:267--305, 2012.

\bibitem{resnet}
Kaiming He, Xiangyu Zhang, Shaoqing Ren, and Jian Sun.
\newblock Identity mappings in deep residual networks.
\newblock In {\em 14th European Conference on Computer Vision}, pages 630--645. Springer, 2016.

\bibitem{Hodan2016}
Tom{\'a}{\v{s}} Hoda{\v{n}}, Ji{\v{r}}{\'\i} Matas, and {\v{S}}t{\v{e}}p{\'a}n Obdr{\v{z}}{\'a}lek.
\newblock On evaluation of 6d object pose estimation.
\newblock In {\em ECCV Workshops}, pages 606--619. Springer, 2016.

\bibitem{Jianwei2022}
Li~Jianwei, Gao Wei, Wu~Yihong, Liu Yangdong, and Shen Yanfei.
\newblock High-quality indoor scene 3d reconstruction with rgb-d cameras: A brief review.
\newblock {\em Computational Visual Media}, 8(3):369--393, 09 2022.

\bibitem{auc}
Yuhe Jin, Dmytro Mishkin, Anastasiia Mishchuk, Jiri Matas, Pascal Fua, Kwang~Moo Yi, and Eduard Trulls.
\newblock Image matching across wide baselines: From paper to practice.
\newblock {\em International Journal of Computer Vision}, 129(2):517–547, October 2020.

\bibitem{Kim2023}
Soohwan Kim and Minkyoung Kim.
\newblock Rotation representations and their conversions.
\newblock {\em IEEE Access}, PP:1--1, 01 2023.

\bibitem{adam}
Diederik~P. Kingma and Jimmy Ba.
\newblock Adam: A method for stochastic optimization, 2017.

\bibitem{Peretroukhin2020}
Valentin Peretroukhin, Matthew Giamou, David~M. Rosen, W.~Nicholas Greene, Nicholas Roy, and Jonathan Kelly.
\newblock A {S}mooth {R}epresentation of {SO(3)} for {D}eep {R}otation {L}earning with {U}ncertainty.
\newblock In {\em Proceedings of {R}obotics: {S}cience and {S}ystems {(RSS'20)}}, Jul. 12--16 2020.

\bibitem{Pitteri2019}
Giorgia Pitteri, Micha{\"e}l Ramamonjisoa, Slobodan Ilic, and Vincent Lepetit.
\newblock On object symmetries and 6d pose estimation from images.
\newblock In {\em 2019 International Conference on 3D Vision (3DV)}, pages 614--622. IEEE, 2019.

\bibitem{kornia}
Edgar Riba, Dmytro Mishkin, Daniel Ponsa, Ethan Rublee, and Gary Bradski.
\newblock Kornia: an open source differentiable computer vision library for pytorch.
\newblock In {\em Proceedings of the IEEE/CVF Winter Conference on Applications of Computer Vision}, pages 3674--3683, 2020.

\bibitem{Voynov2023}
O.~Voynov, G.~Bobrovskikh, P.~Karpyshev, S.~Galochkin, A.~Ardelean, A.~Bozhcnko, E.~Karmanova, P.~Kopanev, Y.~Labutin-Rymsho, R.~Rakhimov, A.~Safin, V.~Serpiva, A.~Artemov, E.~Burnaev, D.~Tsetserukou, and D.~Zorin.
\newblock Multi-sensor large-scale dataset for multi-view 3d reconstruction.
\newblock In {\em 2023 IEEE/CVF Conference on Computer Vision and Pattern Recognition (CVPR)}, pages 21392--21403, Los Alamitos, CA, USA, jun 2023. IEEE Computer Society.

\bibitem{rot3d}
Yi~Zhou, Connelly Barnes, Jingwan Lu, Jimei Yang, and Hao Li.
\newblock On the continuity of rotation representations in neural networks.
\newblock In {\em Proceedings of the IEEE/CVF conference on computer vision and pattern recognition}, pages 5745--5753, 2019.

\end{thebibliography}
